\documentclass{article}
\usepackage{spconf,amsmath,graphicx}

\usepackage{multirow} 
\usepackage{booktabs} 
\usepackage{adjustbox}
\usepackage{amsmath}
\usepackage{amssymb}
\usepackage[table]{xcolor}

\usepackage{xcolor}

\definecolor{rblue}{rgb}{0,0.5,1}
\definecolor{awesome}{rgb}{1.0, 0.13, 0.32}
\definecolor{hollywoodcerise}{rgb}{0.96, 0.0, 0.63}
\definecolor{lasallegreen}{rgb}{0.03, 0.47, 0.19}
\definecolor{hanpurple}{rgb}{0.32, 0.09, 0.98}
\definecolor{green(pigment)}{rgb}{0.0, 0.65, 0.31}
\makeatletter
\let\NAT@parse\undefined
\makeatother
\usepackage[pagebackref=false,breaklinks=true,colorlinks,bookmarks=false]{hyperref}
\hypersetup{colorlinks=true,
    linkcolor={red},
    citecolor={hanpurple},
    urlcolor={magenta}
}

\title{Event-guided 3D Gaussian Splatting for Dynamic Human and Scene Reconstruction}
\name{Xiaoting Yin$^{1}$, Hao Shi$^{1,2}$, Kailun Yang$^{3}$, Jiajun Zhai$^{1}$, Shangwei Guo$^{1}$, Lin Wang$^{2}$, and Kaiwei Wang$^{1,\dag}$ 
\thanks{This work was supported in part by the Natural Science Foundation of Zhejiang Province (Grant No. LZ24F050003), in part by the National Natural Science Foundation of China (NSFC) under Grant No.~12174341 and No.~62473139, in part by the Hunan Provincial Research and Development Project (Grant No. 2025QK3019), and in part by the Open Research Project of the State Key Laboratory of Industrial Control Technology, China (Grant No. ICT2025B20).}%
\thanks{$^{\dag}$Corresponding author.}}
\address{$^{1}$College of Optical Science and Engineering, Zhejiang University, China\\
$^{2}$School of Electrical and Electronic Engineering, Nanyang Technological University, Singapore\\
$^{3}$School of Artificial Intelligence and Robotics, Hunan University, China
}

\begin{document}
\maketitle
\begin{abstract}
Reconstructing dynamic humans together with static scenes from monocular videos remains difficult, especially under fast motion, where RGB frames suffer from motion blur. Event cameras exhibit distinct advantages, \textit{e.g.}, microsecond temporal resolution, making them a superior sensing choice for dynamic human reconstruction. Accordingly, we present a novel event-guided human-scene reconstruction framework that jointly models human and scene from a single monocular event camera via 3D Gaussian Splatting. Specifically, a unified set of 3D Gaussians carries a learnable semantic attribute; only Gaussians classified as human undergo deformation for animation, while scene Gaussians stay static. To combat blur, we propose an event-guided loss that matches simulated brightness changes between consecutive renderings with the event stream, improving local fidelity in fast-moving regions. Our approach removes the need for external human masks and simplifies managing separate Gaussian sets. On two benchmark datasets, ZJU-MoCap-Blur and MMHPSD-Blur, it delivers state-of-the-art human-scene reconstruction, with notable gains over strong baselines in PSNR/SSIM and reduced LPIPS, especially for high-speed subjects. 
\end{abstract}

\begin{figure}[!t]
\centering
\includegraphics[width=0.85\columnwidth]{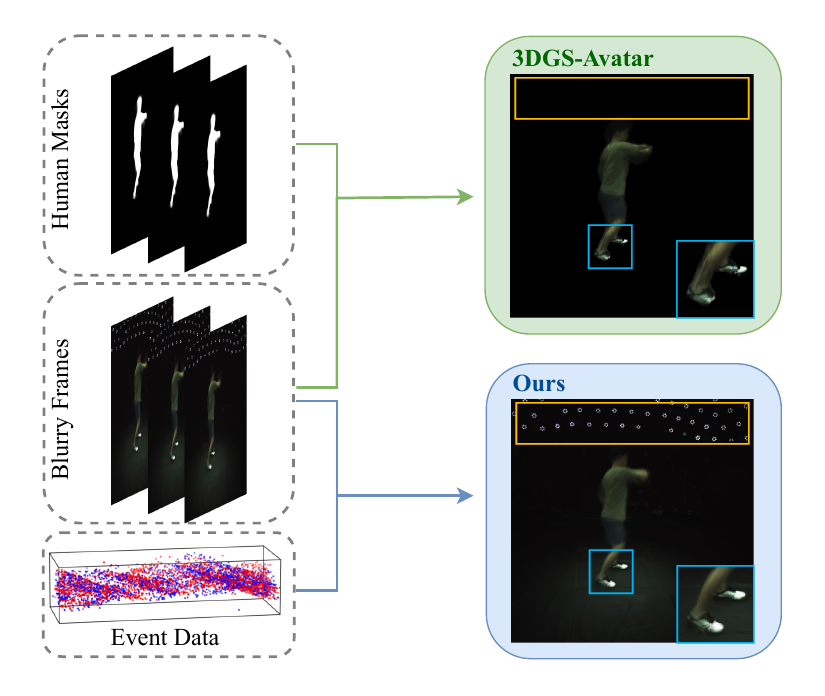}
\vskip -1.5ex
\caption{\textbf{Comparison of the baseline method and our method. }
Our approach jointly reconstructs humans and scenes, leveraging event data to mitigate motion blur.
}
\vskip -1.5ex
\label{fig:intro}
\end{figure}

\begin{figure*}[!t]
\centering
\includegraphics[width=0.9\textwidth]{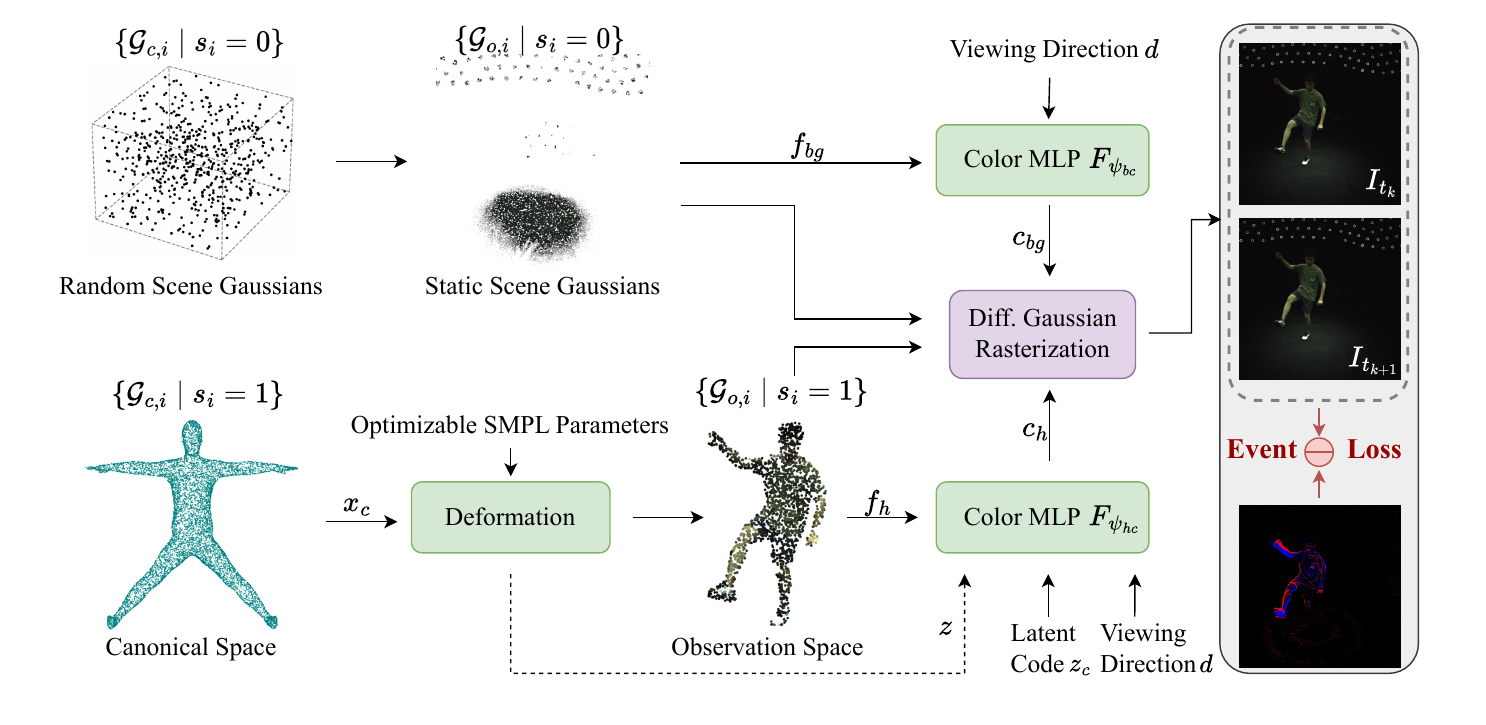}
\caption{\textbf{Overview of our approach. } 
Our framework reconstructs humans and static scenes from a single monocular event camera. 
We model both the human and the scene as a unified set of 3D Gaussians with a semantic attribute. 
The human Gaussians are deformed, while scene Gaussians remain static. 
Leveraging the event camera’s high temporal resolution, we supervise the rendered images with real event data to mitigate motion blur.
}
\label{fig:pipeline}
\end{figure*}

\section{Introduction}
\label{sec:intro}
Human reconstruction from monocular videos is a critical task in computer vision and graphics, with applications spanning virtual reality~\cite{morgenstern2024animatable}, augmented reality~\cite{kyrlitsias2022social}, and film production~\cite{habermann2019livecap}. Recent neural rendering advancements, including Neural Radiance Fields (NeRFs)\cite{mildenhall2021nerf} and 3D Gaussian Splatting (3DGS)\cite{kerbl20233d}, enable highly-fidelity, photorealistic 3D reconstruction. Building on this, various 3D human reconstruction methods have emerged. Examples include 3DGS-Avatar~\cite{qian20243dgs} and ASH~\cite{pang2024ash}, which focus on animatable avatars, and HUGS~\cite{kocabas2024hugs}, which reconstructs human and scene simultaneously using separate Gaussian sets.

Despite these promising results, existing methods still face significant challenges. 
First, most approaches require an external human mask, necessitating a prior segmentation step that can introduce artifacts.
Second, rapid human motion in frame-based camera captures often leads to motion blur, deteriorating image quality.
While some methods attempt to deblur RGB images or integrate event data for reconstruction, their generalizability is limited. 
ExFMan~\cite{chen2024exfman} is a notable exception that leverages event data for dynamic human reconstruction but lacks static scene modeling.

To address these challenges, we introduce a unified framework for reconstructing animatable humans and static scenes from a monocular event camera (Fig.~\ref{fig:intro}). 
Unlike HUGS~\cite{kocabas2024hugs}, which uses separate Gaussian sets, our method encodes both human and scene in a single set of 3D Gaussians with semantic attributes, refined during training via rendering feedback.
Furthermore, synthetic events generated from rendered images are aligned with real event streams, providing supervision that alleviates motion blur.

We evaluate our method on two newly created datasets, ZJU-MoCap-Blur and MMHPSD-Blur, generated by simulating motion blur to test performance under challenging conditions. 
Experiments show that our unified human-scene reconstruction framework surpasses the state-of-the-art HUGS~\cite{kocabas2024hugs}, with notable gains on ZJU-MoCap-Blur: +$19.5\%$ PSNR, +$3.95\%$ SSIM, and –$32.5\%$ LPIPS.
In summary, our main contributions are:
\begin{itemize}
\item A novel framework for unified human and scene reconstruction using a single semantically attributed set of 3D Gaussians.
\item The integration of event data to mitigate motion blur and enhance the reconstruction quality of fast-moving subjects.
\item An extensive evaluation on self-generated motion-blurred datasets that demonstrates state-of-the-art performance in challenging high-speed scenarios.
\end{itemize}

\section{Methodology}
\label{sec:method}

\subsection{Overview}
Our framework reconstructs animatable humans and static scenes from a single monocular event
camera (See Fig.~\ref{fig:pipeline}). 
We first review 3D Gaussian Splatting (3DGS) and the event camera model (Sec.~\ref{method:preliminaries}). 
We then extend 3DGS with semantic attributes for unified human–scene representation (Sec.~\ref{method:semantic_3dgs}), enhance static appearance with a background color MLP (Sec.~\ref{Method:bg_color_mlp}), and apply an event-guided loss to supervise rendering and reduce motion blur (Sec.~\ref{method:event_loss}).

\subsection{Preliminaries}
\label{method:preliminaries}

~~~~~~\textit{1) 3DGS-Avatar: }
3DGS-Avatar~\cite{qian20243dgs} extends 3DGS for animatable human avatars by optimizing a set of 3D Gaussians in a canonical space, where a non-rigid deformation network models subtle changes such as clothing wrinkles:
\begin{equation}
\{\mathcal{G}_d\} = \Phi_{\psi_{nr}}(\{\mathcal{G}_c\}, {z}_p), 
\label{eq:gs-avatar_non_rigid}
\end{equation}
where $\mathbf{z}_p$ is an encoded pose vector. 
These deformed Gaussians are then transformed to the observation space via a rigid transformation to align them with a specified pose, which uses Linear Blend Skinning (LBS)~\cite{loper2023smpl}:
\begin{equation}
\{\mathcal{G}_o\} = \Phi_{\psi_{r}}(\{\mathcal{G}_d\}; \{\mathbf{B}_b\}_{b=1}^B), \label{eq:rigid_def}
\end{equation}
where a skinning MLP $\Phi_{\psi_{r}}$ predicts weights at position $\mathbf{x}_d$, and $\{\mathbf{B}_b\}_{b=1}^B$ are bone transformations. 

\textit{2) Event Camera Model: }
Unlike conventional frame-based cameras that capture intensity images at a fixed rate, event cameras operate asynchronously \cite{zheng2023deep, Rebecq19pami}. 
Each pixel independently reports a brightness change as a discrete event \(e_k = (u_k, t_k, p_k)\), defined by its pixel coordinates \(u_k\), timestamp \(t_k\), and polarity \(p_k\). 
An event is triggered when the logarithmic brightness at a pixel, $L(u_k, t_k) = \log I(u_k, t_k)$, accumulates a change that exceeds a contrast threshold $C$ since the last event at that pixel:
\begin{equation}
L(u_k, t_k) - L(u_k, t_{k-1}) = p_k \cdot C.
\label{eq:event_model}
\end{equation}
This asynchronous, high-temporal-resolution data makes event cameras robust to motion blur and highly suitable for capturing high-speed dynamic scenes.

\subsection{Unified Human-Scene Representation}
\label{method:semantic_3dgs}
To unify the representation of animatable humans and static scenes, our method introduces a semantic property for each 3D Gaussian.
The semantic property $s_i$ for each Gaussian $\mathcal{G}_o$ is initialized based on the available semantic labels $L_i$ for the initial point cloud:
\begin{equation}
{s}_i = \mathcal{L}_i \in {0, 1}.
\label{eq:semantic_init}
\end{equation}
The semantic attribute $s_i$ for a given Gaussian $\mathcal{G}_o$ is a learnable parameter. 
We obtain a soft mask value $m_i$ for each Gaussian using a sigmoid function:
\begin{equation}
{m}_i = \sigma({s}_i).
\label{eq:semantic_mask}
\end{equation}
The soft mask value ${m}_i$ is then binarized using a threshold of $0.5$ to create a hard mask ${s}_i \in \{0, 1\}$. 
Only Gaussians with a hard mask value of $1$ (classified as human) are passed to the deformation networks for animation:
\begin{equation}
\{\mathcal{G}_{o,i}\} = {s}_i \cdot \Phi_{\psi_r}(\Phi_{\psi_{nr}}(\{\mathcal{G}_c\}, {z}_p); \{\mathbf{B}_b\}_{b=1}^B).
\label{eq:conditional_deformation}
\end{equation}
Similarly, the final color for each Gaussian is determined by its classified semantic category, with human and scene Gaussians being processed by separate color MLPs to produce their respective colors ${c}_{h}$ and ${c}_{bg}$.
This semantic property is also integrated into the densification process of 3DGS, as new Gaussians inherit the semantic properties of their parents.

\subsection{Static Scene Appearance Modeling}
\label{Method:bg_color_mlp}
To model the static scene appearance, we employ a dedicated scene color MLP. 
This approach provides a more expressive representation compared to traditional Spherical Harmonics (SH) methods, especially when handling challenging data conditions such as noise or motion blur. 
For each scene Gaussian ($\{ \mathcal{G}_{c,i} \mid s_i = 0 \}$), the MLP takes a learnable feature vector and an SH basis as input to predict its final color.
Specifically, the MLP takes the feature vector $\mathbf{f}_{bg}$ and the SH basis $\gamma({\mathbf{d}})$ of the viewing direction $\mathbf{d}$ as input to predict the scene color ${c}_{bg}$:
\begin{equation}
c_{bg} = \mathcal{F}_{\psi_{bc}}(\mathbf{f}_{bg}, \gamma(\mathbf{d})), 
\label{eq:scene_color_mlp}
\end{equation}
where $\mathcal{F}_{\psi_{bc}}$ is a multi-layer perceptron (MLP). 
This approach combines the expressiveness of a neural network with the directional encoding of SH, enabling robust non-linear color modeling for background regions.

\subsection{Event Loss}
\label{method:event_loss}
To effectively mitigate motion blur caused by fast human movements, we introduce an Event Loss that leverages the high temporal resolution of event data. 
Drawing from the event camera model in Section~\ref{method:preliminaries}, a change in logarithmic brightness triggers an event. 
We simulate this process by calculating the per-pixel logarithmic brightness change between consecutive rendered frames, $I_{t_k}$ and $I_{t_{k+1}}$:
\begin{equation}
\Delta \mathcal{L} = \log(I_{t_{k+1}}^{2.2} + \epsilon) - \log(I_{t_k}^{2.2} + \epsilon),
\label{eq:delta_log_lum}
\end{equation}
where the images are first converted from sRGB to linear space~\cite{iec_standard} by raising them to the power of $2.2$, and $\epsilon$ is a constant to prevent numerical instability. 
The resulting map $\Delta \mathcal{L}$ represents the simulated events, where each pixel's value indicates the magnitude and polarity of the brightness change.

Our final event loss is then formulated as a normalized $L1$ distance between the simulated events $\Delta \mathcal{L}$ and the ground truth event data $E_{gt}$. 
Normalization is applied to ensure the loss is robust to varying light conditions and event densities across different frames. 
The loss is computed as:
\begin{equation}
\mathcal{L}{\text{event}} = w_{ev} \cdot \left| \frac{\Delta \mathcal{L}}{||\Delta \mathcal{L}||_F} - \frac{\mathbf{E}{gt}}{||\mathbf{E}_{gt}||_F} \right|_1,
\label{eq:event_loss}
\end{equation}
where $\| \cdot \|_F$ denotes the Frobenius norm, which is used to normalize the pixel-wise values, and $w_{ev}$ is a weighting factor.  
This loss encourages our model's rendered images to replicate the brightness changes observed by the event camera, improving reconstruction quality in high-speed scenarios.

\begin{figure}[!t]
\centering
\includegraphics[width=\columnwidth]{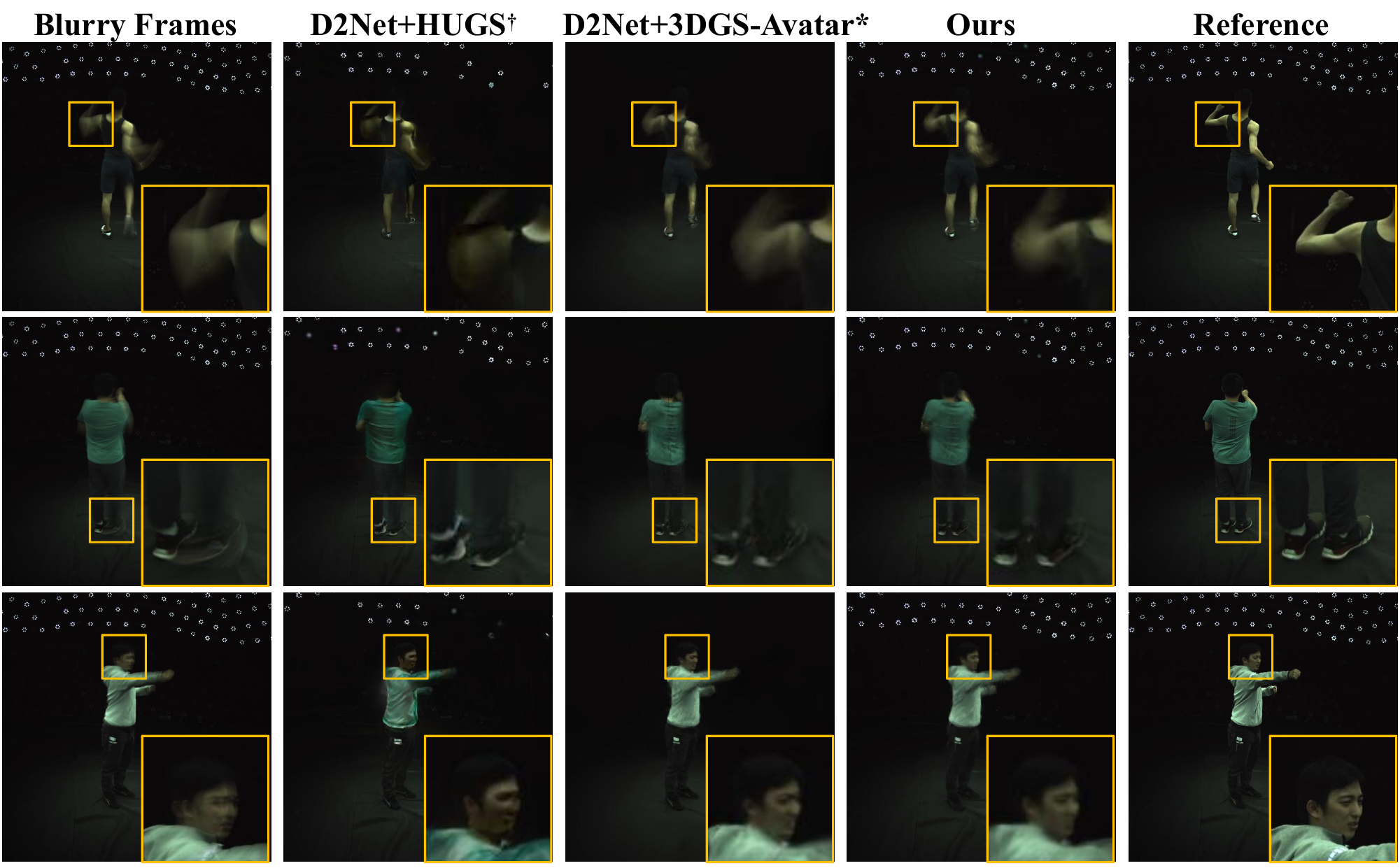}
\caption{\textbf{Qualitative results on ZJU-MoCap-Blur dataset.} 
}
\label{fig:zju}
\end{figure}

\section{Experiments}
\label{sec:experiments}

\subsection{Experimental Settings}
~~~~~~\textit{1) Datasets:}
To simulate blurry images, Super-SloMo~\cite{jiang2018super} is applied to sequences from the ZJU-MoCap~\cite{peng2021neural} and MMHPSD~\cite{zou2021eventhpe} datasets. For ZJU-MoCap-Blur, six sequences (377, 386, 387, 392, 393, 394) from view ``1'' are used, with the last three images out of every ten designated as the test set. 
The MMHPSD-Blur dataset utilizes six sequences (s1g2t3, s5g1t1, s7g1t1, s10g3t4, s14g2t2, s15g3t4). Human masks are generated using RobustVideoMatting~\cite{lin2022robust}.

\textit{2) Baselines \& Metrics:}
We compare our method against two baselines: 3DGS-Avatar~\cite{qian20243dgs} and HUGS~\cite{kocabas2024hugs}. 
3DGS-Avatar is extended to 3DGS-Avatar* for simultaneous human and scene rendering by integrating semantic attributes, initialized with initial values of $0.5$. 
HUGS$^\dagger$ utilizes the official codebase~\cite{kocabas2024hugs}, incorporating random cubic sampling for scene point cloud initialization to ensure fair comparison.
Both 3DGS-Avatar* and HUGS$^\dagger$ are also cascaded with RGB-based deblurring~\cite{Zamir2021MPRNet, chen2022simple}, and RGB+Event-based deblurring~\cite{sun2022event, Shang_2021_ICCV} methods for comprehensive comparison. 
Reconstruction quality is quantitatively evaluated using PSNR, SSIM, and LPIPS.

\begin{figure}[!t]
\centering
\includegraphics[width=\columnwidth]{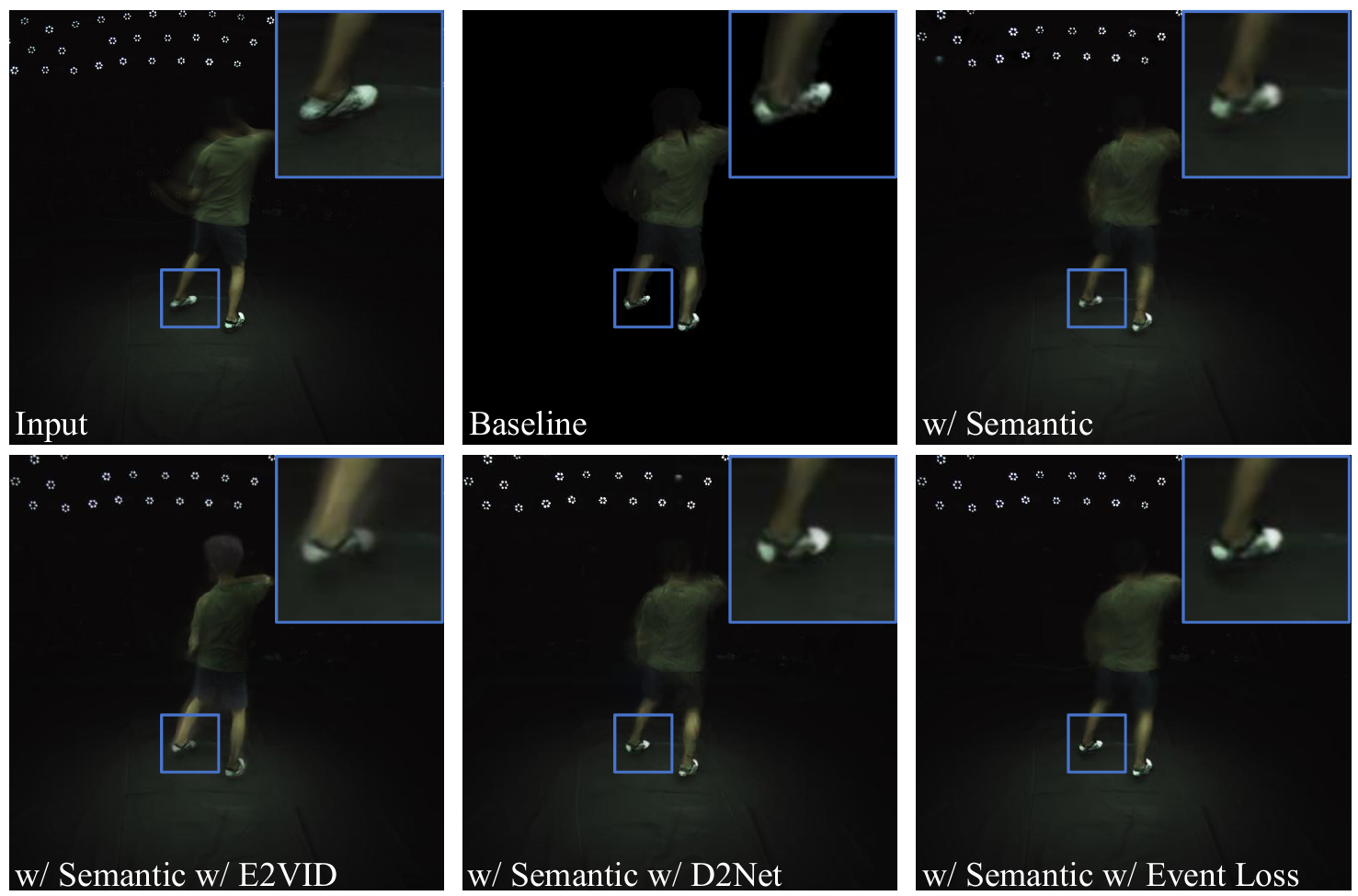}
\caption{\textbf{Visual analysis of ablation study.} }
\label{fig:ablation}
\end{figure}

\begin{table}[!t]
\centering
\vskip -2ex
\caption{\textbf{Quantitative comparison on ZJU-MoCap-Blur dataset.} 
\textbf{Bold} numbers represent the best and \underline{underlined} numbers represent the second-best.}
\label{tab:comp_zju}
\adjustbox{max width=\columnwidth}{
\begin{tabular}{l l c c c}
\toprule
\multirow{2}{*}{\textbf{Category}} & \multirow{2}{*}{\textbf{Method}} & \multicolumn{3}{c}{\textbf{Metrics}} \\
\cmidrule(lr){3-5}
 & & \textbf{PSNR} & \textbf{SSIM} & \textbf{LPIPS} \\
\midrule
\midrule
\multirow{3}{*}{Baselines} 
 & 3DGS-Avatar~\cite{qian20243dgs} & 21.75 & 0.2042 & 0.4026 \\
 & 3DGS-Avatar*~\cite{qian20243dgs} & 25.47 & 0.9441 & 0.1917 \\
 & HUGS$^\dagger$~\cite{kocabas2024hugs} & 26.70 & 0.9366 & 0.1075 \\
\midrule
\multirow{4}{*}{RGB-based Deblur} 
 & MPR~\cite{Zamir2021MPRNet} + 3DGS-Avatar*~\cite{qian20243dgs} & 25.46 & 0.9438 & 0.1912 \\
 & MPR~\cite{Zamir2021MPRNet} + HUGS$^\dagger$~\cite{kocabas2024hugs} & 26.87 & 0.9382 & \underline{0.1042} \\
 & NAFNet~\cite{chen2022simple} + 3DGS-Avatar*~\cite{qian20243dgs} & 25.47 & 0.9443 & 0.1902 \\
 & NAFNet~\cite{chen2022simple} + HUGS$^\dagger$~\cite{kocabas2024hugs} & 26.35 & 0.9327 & 0.1189 \\
\midrule
\multirow{4}{*}{RGB+Event Deblur} 
 & EFNet~\cite{sun2022event} + 3DGS-Avatar*~\cite{qian20243dgs} & 25.42 & 0.9404 & 0.1929 \\
 & EFNet~\cite{sun2022event} + HUGS$^\dagger$~\cite{kocabas2024hugs} & 26.14 & 0.9277 & 0.1226 \\
 & D2Net~\cite{Shang_2021_ICCV} + 3DGS-Avatar*~\cite{qian20243dgs} & 25.51 & \underline{0.9461} & 0.1909 \\
 & D2Net~\cite{Shang_2021_ICCV} + HUGS$^\dagger$~\cite{kocabas2024hugs} & \underline{26.93} & 0.9391 & 0.1080 \\
\midrule
 \rowcolor{gray!20}& Ours & \textbf{31.91} & \textbf{0.9736} & \textbf{0.0726} \\
\bottomrule
\end{tabular}
}
\vskip -2ex
\end{table}

\subsection{Comparisons}

~~~~~~\textit{1) Qualitative: }Fig.~\ref{fig:zju} illustrates that compared methods struggle with background reconstruction and produce incomplete scenes on the ZJU-Mocap-Blur dataset, whereas our method robustly achieves simultaneous human and scene reconstruction, effectively mitigating dynamic blur from the input blurry images.

\textit{2) Quantitative: }Tables~\ref{tab:comp_zju} and~\ref{tab:comp_mmhpsd} present the quantitative results, demonstrating our method's superior performance (higher PSNR/SSIM) on both the ZJU-Mocap-Blur and MMHPSD-Blur datasets compared to baseline methods and their cascaded deblurring extensions.

\subsection{Ablation Study}
We ablate our method on sequence $392$ of the ZJU-Mocap-Blur dataset, with results reported in Tab.~\ref{tab:ablation1_attributes} and Fig.~\ref{fig:ablation}.
Initially, extending the 3DGS-Avatar~\cite{qian20243dgs} baseline with semantic attributes enables simultaneous human and scene reconstruction. 
While cascading with the RGB+Event deblurring method~\cite{Shang_2021_ICCV} improves image quality by reducing motion blur, incorporating an event loss supervision further enhances image fidelity.

\begin{table}[!t]
\centering
\caption{\textbf{Quantitative results on MMHPSD-Blur dataset.}}
\label{tab:comp_mmhpsd}
\adjustbox{max width=0.7\columnwidth}{
\begin{tabular}{lccc}
\toprule
\multirow{2}{*}{\textbf{Method}} & \multicolumn{3}{c}{\textbf{Metrics}} \\
\cmidrule(lr){2-4}
 & \textbf{PSNR} & \textbf{SSIM} & \textbf{LPIPS} \\
\midrule
\midrule
3DGS-Avatar~\cite{qian20243dgs} & 6.86 & 0.3667 & 0.4956 \\
3DGS-Avatar*~\cite{qian20243dgs} & 15.15 & 0.7335 & 0.4205 \\
HUGS$^\dagger$~\cite{kocabas2024hugs} & \underline{25.23} & \underline{0.8447} & \textbf{0.1167} \\
MPR~\cite{Zamir2021MPRNet} + HUGS$^\dagger$~\cite{kocabas2024hugs} & 25.07 & 0.8405 & 0.1213 \\
NAFNet~\cite{chen2022simple}  + HUGS$^\dagger$~\cite{kocabas2024hugs} & 24.93 & 0.8415 & \underline{0.1183}  \\
D2Net~\cite{Shang_2021_ICCV} + HUGS$^\dagger$~\cite{kocabas2024hugs} & 24.92 & 0.8324 & 0.2153 \\
\rowcolor{gray!20}Ours & \textbf{25.91} & \textbf{0.9118} & 0.1321 \\
\bottomrule
\end{tabular}
}
\end{table}

\begin{table}[t]
\centering
\caption{\textbf{Ablation study of semantic attributes and event incorporation.}}
\label{tab:ablation1_attributes}
\adjustbox{max width=\columnwidth}{
\begin{tabular}{lccccc}
\toprule
\textbf{Method} & \textbf{Semantic} & \textbf{Event} & \textbf{PSNR} & \textbf{SSIM} & \textbf{LPIPS} \\
\midrule
\midrule
Baseline (3DGS-Avatar~\cite{qian20243dgs}) & $\times$ & $\times$ & 21.68 & 0.2076 & 0.4129 \\
+ Semantic & $\checkmark$ & $\times$ & 31.15 & 0.9716 & 0.0876 \\
+ Sem. + E2VID~\cite{Rebecq19pami, rebecq2019events} & $\checkmark$ & $\checkmark$ & 30.25 & 0.9693 & 0.0799 \\
+ Sem. + RGB\&Event Deblur~\cite{Shang_2021_ICCV}  & $\checkmark$ & $\checkmark$ & 31.44 & 0.9738 & 0.0921\\
\rowcolor{gray!20}+ Sem. + Event Loss & $\checkmark$ & $\checkmark$ & \textbf{31.81} & \textbf{0.9730} & \textbf{0.0813} \\
\bottomrule
\end{tabular}
}
\end{table}

\section{Conclusion}
\label{sec:conclusion}
In this paper, we have presented a unified event-aided 3D Gaussian Splatting framework that reconstructs dynamic humans and static scenes from a single monocular event camera. By assigning a learnable semantic attribute to each Gaussian and introducing an event-driven supervision loss, our method removes the need for external human masks and robustly mitigates motion blur. 
Across two motion-blur benchmarks, it delivers state-of-the-art human-scene reconstruction with clear gains in fidelity (higher PSNR/SSIM) and visibly reduced motion ghosting, supported by ablations showing complementary benefits from semantic unification and event guidance.

%
\clearpage
\bibliographystyle{IEEEbib}
\bibliography{strings,refs}

\end{document}